\documentclass[]{arxiv_article}

\usepackage[toc,page,header]{appendix}
\microtypesetup{expansion=false}
\usepackage{xspace}
\usepackage{enumitem}
\usepackage[linesnumbered,ruled,vlined]{algorithm2e}
\usepackage{amsmath}
\usepackage{amssymb}
\definecolor{darkgreen}{rgb}{0.0,0.45,0.0}

\newcommand{\keywords}[1]{\par\medskip\noindent\textbf{Keywords:}\space\def\and{, }#1}

\title{Hi-WM: Human-in-the-World-Model for \\ Scalable Robot Post-Training}

\author[1,*]{Yaxuan Li}{}
\author[1,*]{Zhongyi Zhou}{}
\author[1,*]{Yefei Chen}{}

\author[2]{Yanjiang Guo}{}
\author[3]{Jiaming Liu}{}
\author[3]{Shanghang Zhang}{}
\author[2]{Jianyu Chen}{}
\author[4]{Yichen Zhu}{}

\affiliation[1]{Current Robotics}
\affiliation[2]{Tsinghua University}
\affiliation[3]{Peking University}
\affiliation[4]{University of Toronto}

\contribution[*]{Equal Contribution}

\abstract{
Post-training is essential for turning pretrained generalist robot policies into reliable task-specific controllers, but existing human-in-the-loop pipelines remain tied to physical execution: each correction requires robot time, scene setup, resets, and operator supervision in the real world. Meanwhile, action-conditioned world models have been studied mainly for imagination, synthetic data generation, and policy evaluation. We propose \textbf{Human-in-the-World-Model (Hi-WM)}, a post-training framework that uses a learned world model as a reusable corrective substrate for failure-targeted policy improvement. A policy is first rolled out in closed loop inside the world model; when the rollout becomes incorrect or failure-prone, a human intervenes directly in the model to provide short corrective actions. Hi-WM caches intermediate states and supports rollback and branching, allowing a single failure state to be reused for multiple corrective continuations and yielding dense supervision around behaviors that the base policy handles poorly. The resulting corrective trajectories are then added back to the training set for post-training. We evaluate Hi-WM on three real-world manipulation tasks spanning both rigid and deformable object interaction, and on two policy backbones. Hi-WM improves real-world success by 37.9 points on average over the base policy and by 19.0 points over a world-model closed-loop baseline, while world-model evaluation correlates strongly with real-world performance (r = 0.953). These results suggest that world models can serve not only as generators or evaluators, but also as effective corrective substrates for scalable robot post-training.

}

\date{\today}
\checkdata[Project Page]{\url{https://hi-wm.github.io/}}

\teaser{
    \begin{center}
        \includegraphics[width=0.96\textwidth]{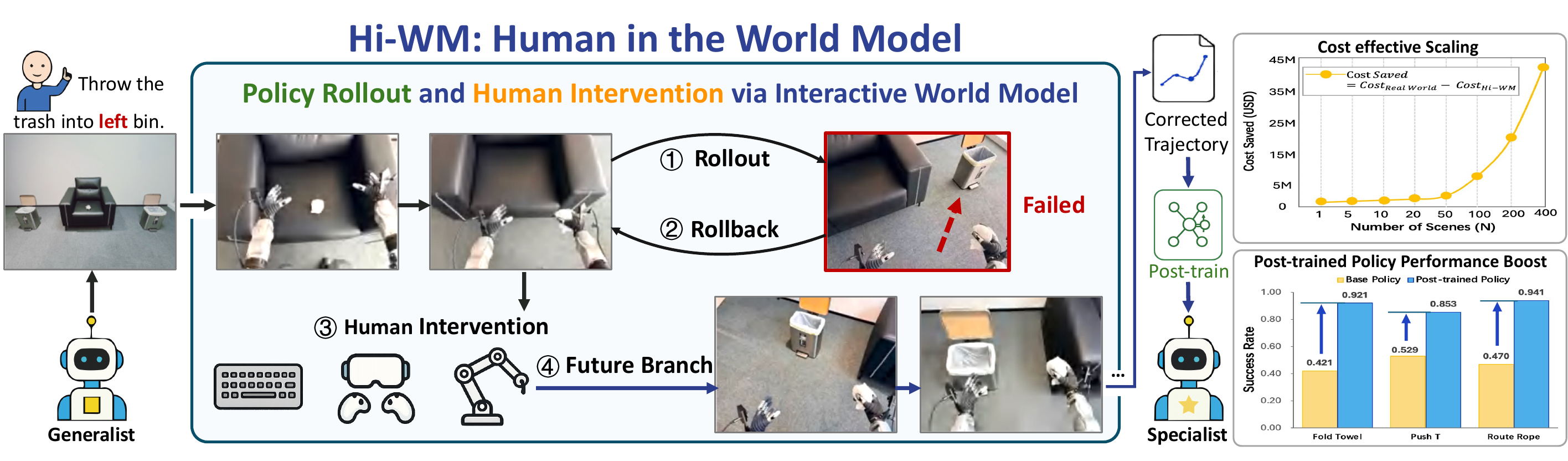} 
        \captionof{figure}{Hi-WM overview. A policy is rolled out in a learned world model. When the rollout becomes failure-prone, a human provides short corrective actions. Cached states support rollback and branching, producing multiple continuations for post-training.} 
        \label{fig:firstfig}
    \end{center}
}

\begin{document}
\begingroup
\hypersetup{linkcolor=black}
\maketitle
\endgroup

\section{Introduction}
Large-scale robot pretraining has produced generalist policies that transfer across tasks, objects, and language instructions. Yet broad pretraining alone is rarely sufficient for deployment. In concrete scenes, policies still fail under contact-rich dynamics, environment-specific variation, and long-horizon error accumulation. In practice, post-training is what turns a generalist policy into a reliably deployable, task- or scene-specialized policy.

The most effective post-training pipelines are still on-policy and corrective: let the current policy act, observe where it breaks, collect targeted supervision on those failure states, and update the policy accordingly [refs]. However, this loop remains tied to the physical world. Every rollout requires real robots, real scenes, human monitoring, and often manual reset. As coverage expands to more tasks, scenes, checkpoints, and long-tail failure modes, the marginal cost of post-training grows rapidly. For generalist policies, the bottleneck is therefore not only data efficiency, but deployment economics.

World models suggest a different substrate for post-training. Prior work has used learned dynamics primarily for imagination, synthetic data generation, policy optimization, or post-hoc evaluation. In contrast, we study the world model as an interactive corrective workspace for post-training. Our key observation is that the most valuable supervision often lies near the learner’s own failure states: not in generic successful rollouts, but in short corrective continuations that recover from imminent failure. If such corrections are collected only in the physical world, each intervention remains tied to hardware time, scene reset, and sequential deployment.

We therefore propose \textbf{Human-in-the-World-Model (Hi-WM)}, a post-training framework in which a pretrained policy is rolled out inside an action-conditioned world model, and a human intervenes only when the rollout becomes incorrect, uncertain, or failure-prone. Instead of treating the world model as a generator or evaluator alone, Hi-WM uses it as a reusable corrective substrate: failure states can be cached, rewound, and branched into multiple alternative continuations, yielding dense recovery supervision without repeated real-world execution. The resulting corrective trajectories are then used for post-training, e.g., with imitation learning or reinforcement learning. In this way, Hi-WM preserves the benefits of targeted human correction while changing the cost structure of post-training from a hardware-bound loop to a reusable interactive process.

We instantiate Hi-WM on three real-world manipulation tasks spanning rigid and deformable object interaction. Across these tasks, we study whether the world model is faithful enough to support intervention, whether world-model-collected corrective trajectories improve real-world policy performance, and whether the benefits scale with additional intervention data that are collected from the world simulator. Our results show that Hi-WM consistently improves real-world success over both the base policy and a world-model closed-loop baseline, suggesting that the main benefit comes not simply from more simulated rollouts, but from failure-targeted human correction inside the learned environment.

Our contributions are three-fold:
\begin{itemize}
    \item \textbf{Human corrective post-training in the world model.} We formulate post-training as targeted human correction inside an action-conditioned world model, rather than collecting corrective supervision only during physical robot execution.
    \item \textbf{Rollback-and-branching corrective data collection.} We introduce a data collection mechanism that caches pre-failure states and reuses them to collect multiple corrective continuations, increasing the density and diversity of recovery supervision around failure-prone states.
    \item \textbf{Real-robot evidence for effectiveness and scaling.} We show on three real-world manipulation tasks and two policy backbones that corrective trajectories collected in the world model improve downstream real-world performance, while offering increasingly favorable economics as virtual intervention data scale.
\end{itemize}

\section{Related Work}

\noindent\textbf{Post-training with human corrective supervision.} Human-in-the-loop robot learning has long studied how to improve policies under the learner-induced state distribution by querying expert correction during deployment or interactive training~\cite{liu2023sirius,jiang2024transic,wu2025robocopilot,Mandlekar2023,Welte2025}. More recent work extends this idea to generalist robot policies and post-deployment refinement, where human interventions, preferences, or corrective demonstrations are used to adapt a pretrained policy to downstream tasks~\cite{ross2011dagger,laskey2017dart,kelly2019hgdagger,spencer2020interventions,jauhri2021tips,hoque2022thriftydagger,hoque2023fleetdagger, luo2024hilserl, luo2024serl}. These approaches show the value of targeted supervision near policy failure. However, the corrective loop remains predominantly tied to physical execution: each intervention consumes robot time, scene setup, reset effort, and operator attention in the real world. Our work addresses this bottleneck by moving corrective interaction from the robot to a learned environment.

\noindent \textbf{World models for data generation and policy improvement.}
A separate line of work uses world models~\cite{wu2023daydreamer,seo2023mwm,mendonca2023structuredwm} as scalable sources of training signal, including imagination for policy learning, synthetic trajectory generation~\cite{koh2021pathdreamer,bruce2024genie,bar2024navigation,awiszus2021worldgen,zhu2024maniwm,seo2023multi,ren2023surfer,zhang2024pivot,li2025robotic,huang2025enerverse,jiang2025enerverse,wang2026interactive,ali2025world,sharma2026world,guo2026vlaw,xiao2025world,jiang2026wovr,chi2025wow,liu2026world,guo2025ctrl,jia2025video2act}, and world-model-based policy optimization~\cite{bharadhwaj2024gen2act,qiu2025lucibot,jang2025dreamgen,hu2024video,chen2024igor,ye2024latent,guo2024prediction,du2023learning,black2023zero,guo2026vlaw,sharma2026world,zhou2025act2goal,ali2025world,chi2025wow,guo2025ctrl,zhang2026veo}. These methods leverage learned dynamics to expand coverage beyond manually collected demonstrations and reduce dependence on expensive real-world interaction. Despite their differences, the world model is typically used as an \textbf{autonomous generator, simulator, or optimizer}. By contrast, our focus is not autonomous rollout alone, but \textbf{human-guided corrective interaction inside the world model}, with the resulting trajectories used specifically for post-training around failure-prone states.

\noindent \textbf{World model for policy evaluator.} World models are also increasingly studied as evaluation~\cite{li2025simpler,zhou2025autoeval,gao2025stargen} substrates for robot policies, where the central question is whether model rollouts preserve policy ranking, correlate with real-world success, and support tasks such as checkpoint selection or pre-deployment screening~\cite{li2025worldeval,quevedo2025worldgym,guo2025ctrl,1xWorldModel,wang2026interactive,ho20251x,huang2025enerverse,jiang2025enerverse,ali2025world,guo2026vlaw,sharma2026world,jiang2025world4rl,liao2025genie,team2025evaluating}. This literature highlights that useful evaluation environments require not only visual realism, but also action faithfulness and consistency with real-world outcomes. Our work is complementary: rather than using the world model only as a post-hoc evaluator, we use it as an interactive corrective workspace in which evaluation, intervention, and corrective data collection are coupled within the same learned environment.

In summary, prior work has explored human corrective learning in the physical world, and world models as generators, evaluators, or optimization environments. Hi-WM lies at the intersection of these directions but addresses a distinct problem setting: \textbf{how to use a world model as a reusable workspace for human-in-the-loop post-training}. The defining feature of our formulation is that failure states inside the learned environment can be cached, rewound, and branched into multiple corrective continuations, converting one-shot physical correction into reusable recovery supervision for post-training.

\section{Methodology}

\subsection{The Human-in-the-World-Model Paradigm}
We formalize Human-in-the-World-Model (Hi-WM) as a novel post-training paradigm that shifts the human intervention loop from physical execution to an interactive world model. Unlike standard physical deployment, which is constrained by sequential and hardware-bound data collection, the Hi-WM paradigm has three practical properties that make this post-training loop scalable:
\begin{itemize}
\item \textbf{State caching and trajectory branching.} 
Because the rollout happens inside the world model, intermediate states can be cached throughout the interaction.
This allows for a direct rollback to any previously failed timestep, from which the system can execute new actions to form alternative recovery branches.

\item \textbf{Reset-free data collection. } Achieving reset-free operation is a crucial prerequisite for fully automated evaluation, yet it remains a formidable challenge in physical settings. By virtualizing the interaction environment, this approach removes the need for manual scene resets, naturally enabling continuous corrective interaction.

\item \textbf{Hardware-agnostic remote collection.} Since human operators interact exclusively with the world model, the data collection process is completely separated from the physical robot. This allows researchers to gather corrective data remotely and in parallel, without being restricted by hardware access or location.
\end{itemize}
Ultimately, these mechanisms transform the human intervention loop from a heavily constrained physical process into a highly scalable pipeline.

\begin{figure}
    \centering
    \includegraphics[width=1\linewidth]{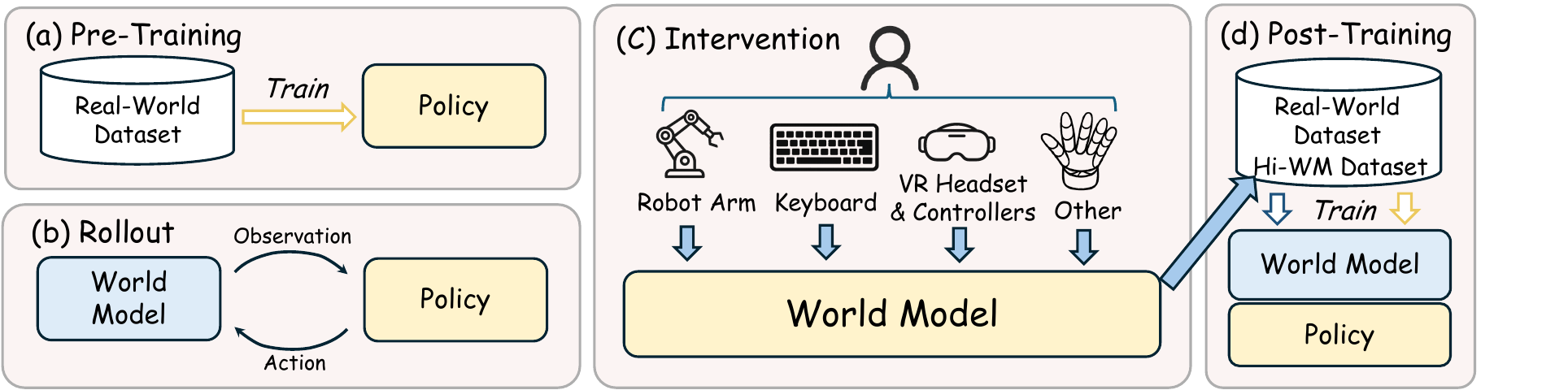}
    \vspace{-1em}
    \caption{\textbf{Human-in-the-World-Model (Hi-WM) pipeline. }
(a) In the pre-training stage, the policy network is learned from real-world data. (b) The policy is rolled out in closed-loop inside the world model. (c) When the policy exhibits suboptimal behavior, a human operator intervenes through hardware-agnostic input devices, such as a robot arm, keyboard, or VR controllers, and the corrective actions are executed directly in the world model. (d) The collected intervention segments are then added to the training set for post-training.}
    \label{fig:framework}
\end{figure}

\subsection{System Overview and Design Requirements}

Hi-WM is built around three components: an action-conditioned world model, a robot policy, and a hardware-agnostic human intervention interface. Together, these components form a closed-loop system for policy rollout, human intervention, and post-training data collection. During rollout, the policy takes the current observation as input and predicts an action. Instead of executing this action on the physical robot, Hi-WM applies it inside the world model, which then produces the next observation. By repeating this process, the policy can be evaluated continuously in the model. When the rollout begins to deviate from the desired behavior, a human operator can step in and provide corrective actions through the intervention interface.

This system design places three requirements on Hi-WM. First, the world model must respond faithfully to actions and remain reliable not only as task progresses, but also in failure-prone states, since these are the situations in which human intervention is most needed. Second, the intervention interface must convert different input devices into the same continuous action space used by the policy, so that control can move smoothly between autonomous rollout and human intervention. 
Third, the system should support collecting corrective segments in a form that can be directly used for post-training. In practice, this includes revisiting earlier states, branching alternative corrections, and adding the resulting trajectories back to the training set.

The following subsections describe these three parts of the system in turn. Section~\ref{subsec:method_wm} focuses on the interactive world model and the design choices needed to make it respond faithfully to actions. Section~\ref{subsec:method_interface} describes the hardware-agnostic intervention interface. Section~\ref{subsec:method_posttrain} explains how corrected rollouts are collected in closed-loop and used for policy post-training.

\subsection{Interactive World Model for Hi-WM}
\label{subsec:method_wm}
To serve as a reliable interactive simulator in the Hi-WM paradigm, the world model should accurately follow robot actions, reflect object interactions, and support data collection for post-training.
Our model consists of a visual encoder, an action-conditioned latent dynamics model, and a visual decoder. 
We train this model from scratch using three key strategies:

\textbf{High-dimensional action conditioning.} 
In domains such as simulated game environments or basic navigation, world models often simplify the control space by projecting actions onto a low-dimensional plane (e.g., using discrete directional commands like up, down, left, and right). While this simplification effectively ensures stable generation in those contexts, it inherently sacrifices the precise spatial control required for complex robotic manipulation. To meet the rigorous precision requirements of our tasks, the world model should process complete, high-dimensional action inputs. Therefore, we train our model conditioned directly on a 14-dimensional continuous action space. This formulation accommodates a dual-arm setup, where each arm contributes a 6-DoF end-effector pose and a 1-DoF gripper state.

\textbf{Fine-grained action discrimination through failure data.}
To accurately simulate the interaction between objects and robot arms, the world model should follow the commanded action closely, rather than only capturing the overall motion trend (e.g., simply approaching the object).
Relying on such approximations often leads to over-optimistic predictions. For instance, if a robotic manipulator moves generally toward a target, the model might bypass exact action execution and hallucinate a successful interaction, even if the specific orientation is incorrect. 
To prevent the model from relying only on coarse motion trends, we include failure cases and off-task states alongside successful trajectories. Specifically, the dataset contains incomplete tasks, failed grasps, unintended object contacts, and misaligned robot poses. This helps the model distinguish between actions that lead to successful interactions and actions that lead to failure.

\textbf{Edge-case data coverage.}
To make the collected data useful for post-training, the world model must stay well aligned with real-world execution. This correspondence ensures that trajectories collected via the model remain physically meaningful when transferred to the real system.

Specifically, once a user performs an action in the world model during human interaction, the action should be faithfully reproduced in the physical world. If the world model differs too much from real execution in robot motion or object interaction, the collected trajectories become unreliable for post-training.
To address this, our solution is to collect a diverse dataset of edge cases. 
These cases include states near the boundary of the workspace and tasks that require precise gripper alignment.
By capturing these edge-case trajectories, the model can accurately reflect the geometry of the scene and the outcomes of executed actions.
We evaluate this alignment through repeated positioning accuracy in Section~ref{subsubsec:repeated acc} and Figure~\ref{fig:relocate}.

With these designs and data strategies, our world model provides a real-time simulation environment that responds faithfully to actions.
It serves as the core infrastructure where policy rollout, human intervention, and corrective trajectory collection seamlessly integrate to enable scalable post-training.

\subsection{Hardware-Agnostic Intervention Interface}
\label{subsec:method_interface}
To support practical human intervention within the world model, we develop a hardware-agnostic teleoperation interface. 
The goal is to decouple the input device from the target robot while keeping control accurate.

The interface supports multiple input devices to suit different task requirements. For simple tasks, a keyboard provides basic, low-dimensional control. For high-precision corrections, 
operators can use calibrated robot arms to specify end-effector motions, which are then mapped into the policy action space.
Alternatively, VR controllers can be used to map tracked spatial poses directly to the robot end-effectors. Regardless of the hardware used, a unified action mapper converts these varied inputs into the standard continuous action representation required by the policy.

This flexibility avoids reliance on a single setup, making human intervention significantly more practical and efficient.

\begin{figure}
    \centering
    \includegraphics[width=1\linewidth]{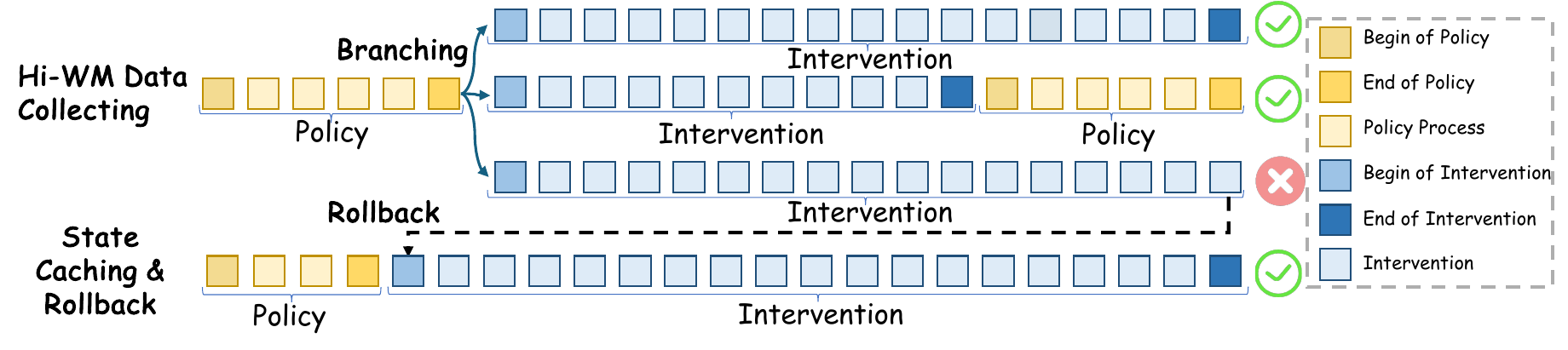}
    \vspace{-1em}
    \caption{\textbf{State caching and rollback mechanism of Hi-WM. }
Hi-WM supports state caching and rollback, allowing a rollout to be rewound to an earlier state and reused to collect multiple corrective branches, thereby improving both the efficiency and diversity of post-training data.}
    \label{fig:rollback}
\end{figure}
\subsection{Collecting Corrective Trajectories for Post-Training}
\label{subsec:method_posttrain}

Within Hi-WM, post-training data are collected in a closed-loop inside the interactive world model. The pre-trained policy first runs in the world model from the current observation and generates a rollout. When the rollout enters unfamiliar or failure-prone states, a human operator intervenes through the hardware-agnostic interface and provides corrective actions. The world model then continues the rollout from the current state using these human actions. Once the rollout has been guided back toward successful behavior, control can be returned to the policy, as shown in Figure ~\ref{fig:rollback}.

A key advantage of operating in the world model is that the system can cache intermediate states and revisit them when needed. Instead of restarting from the beginning after each failure, the operator can rewind the rollout to an earlier timestep and try a different correction from the same state. Consequently, a single simulated failure can be branched into diverse alternative futures, systematically yielding a high-density set of corrective trajectories. 
This makes data collection more efficient and also increases the diversity of recovery behaviors included in the post-training data.

The collected corrective segments are then merged with the original real-world demonstrations and used to post-train the base policy. Because these segments are collected around failure-prone states, they provide targeted supervision on behaviors that the base policy does not handle well. Since this process takes place in the world model rather than on the physical robot, it reduces repeated deployment, manual resets, and continuous hardware use during intervention. 
In this way, Hi-WM turns corrective data collection for post-training from a hardware-bound process into a more efficient and scalable pipeline.
\section{Experiments}

In this section, we conduct extensive real-robot experiments to demonstrate the effectiveness and scalability of our proposed method. Specifically, we aim to answer the following core research questions:

\begin{itemize}
    \item \textbf{RQ1 (Real-World Alignment of Hi-WM):} Is Hi-WM sufficiently aligned with real-world execution to support reliable human intervention and data collection?
    \item \textbf{RQ2 (Effectiveness):} Can Hi-WM serve as an effective, policy-agnostic post-training framework?
    \item \textbf{RQ3 (Scalability):} How does the cost advantage of Hi-WM scale with scene coverage? How does real-world policy performance scale with increasing virtual intervention data?
    \item \textbf{RQ4 (Real-World Generalization):} Does Hi-WM intervention on novel scenarios enhance the policy's real-world generalization?
\end{itemize}

\subsection{Experimental Setup}

\textbf{Real robot tasks.} We evaluate our method on three representative manipulation tasks that cover both rigid and deformable object interaction:
\begin{itemize}
    \item \textbf{Fold Towel.} A towel is placed on the table, and the robot is required to grasp two corners and fold it into a target configuration. This task evaluates coordinated manipulation of deformable objects, requiring accurate corner localization and stable execution during folding.

    \item \textbf{Push-T.} A green T-shaped object is placed on a tabletop with a yellow T-shaped target marker. The robot must push and rotate the object to align it with the target pose. This task evaluates precise pose perception and fine-grained planar manipulation.

    \item \textbf{Route Rope.} A rope is randomly placed on the table, and the robot must grasp its two ends, lift them, and place them into two designated slots so that the rope reaches the desired configuration. This task evaluates grasp pose selection and manipulation of flexible objects under large shape variation.
\end{itemize}

\textbf{Robot setup.} We conduct real-world experiments on a dual-arm YAM-Ultra platform. Each arm provides 6-DoF motion control and a gripper state. Observations are captured from a top-view camera, and robot actions are collected and executed at 15~Hz.

\textbf{Policy eval initialization.} At the beginning of each evaluation rollout, the world model is initialized with a real-world task-start observation, consisting of the initial top-view image of the tabletop scene and the initial poses of the two robot arms. The policy takes the same visual observation and end-effector poses as inputs to generate the first action. The predicted action is then converted into the corresponding end-effector poses and fed to the world model to generate the next observation, which is then used by the policy to produce the subsequent action. In this way, policy evaluation is performed in a closed-loop manner through iterative interaction between the policy and the world model.

\begin{figure}
    \centering
    \includegraphics[width=1\linewidth]{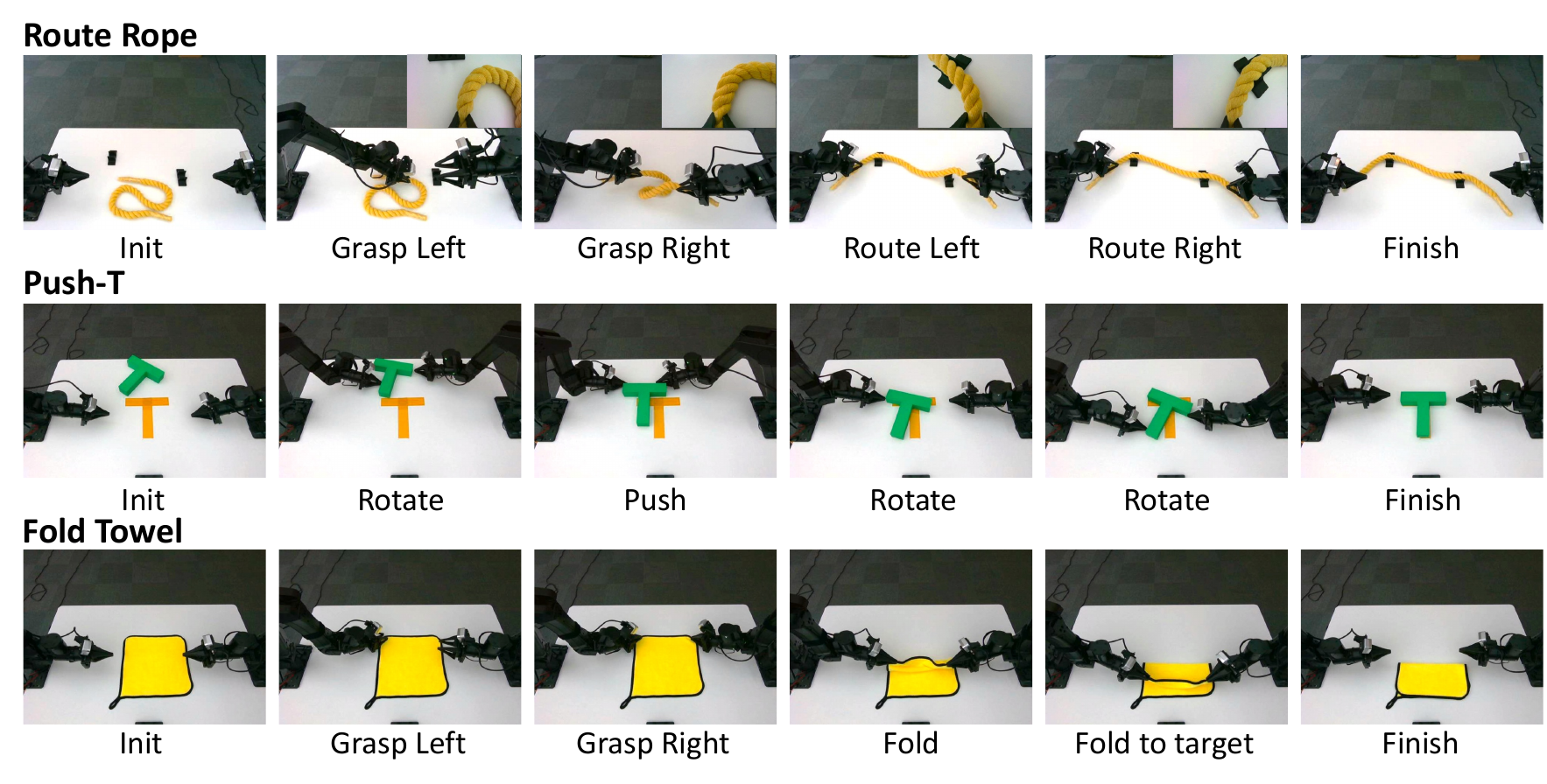}
    \vspace{-1em}
    \caption{\textbf{Evaluated Tasks. }We evaluate our method on three representative manipulation tasks that cover both rigid and deformable object interaction.}
    
    \label{fig:task}
\end{figure}

\textbf{Baselines.} To rigorously evaluate our approach, we compare it against the following baselines:
\begin{itemize}
    \item \textbf{Base Policy}: A policy trained solely on the initial offline dataset without any human intervention.

    \item \textbf{World Model Closed-Loop}: The policy is rolled out within the world model and only the generated successful trajectories are collected. The policy is then fine-tuned on these self-generated data, making this a human-free baseline.

\end{itemize}

\label{subsec:exp_fidelity}
\begin{figure}
    \centering
    \includegraphics[width=0.96\linewidth]{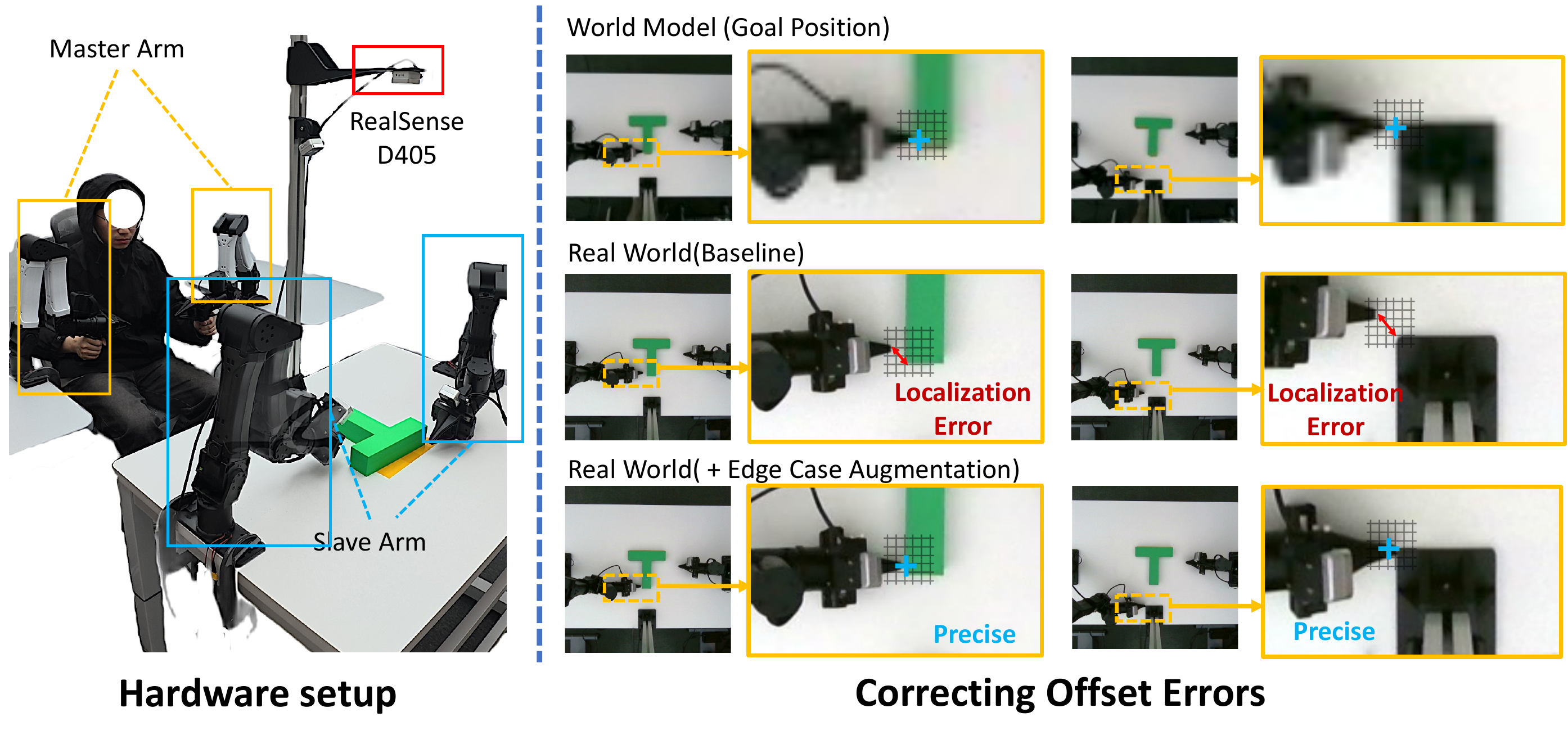}
    \vspace{-1em}
    \caption{\textbf{Hardware setup and offset correction evaluation.} \textbf{Left:} The teleoperation system consisting of a master-slave arm setup and an overhead RealSense D405 camera. \textbf{Right:} Comparison of manipulation precision in the real world. The inclusion of edge-case augmentation corrects significant localization deviations from the world model's target position, closely aligning physical execution with predicted outcomes.}

    \label{fig:relocate}
\end{figure}

\subsection{Real-World Alignment of Hi-WM}

To make human intervention effective in the real world, and to ensure the quality of post-training data, an interactive world model must stay well aligned with physical execution. Visual plausibility alone is not enough. The model must also produce outcomes that can be realized by the real robot. We study this alignment from three perspectives. First, we test whether target positions specified in the world model can be reproduced by the real robot arm. Second, we test whether the model can interpolate to unseen interior points within the data-covered workspace, while remaining stable over long interaction sequences. Third, we test whether task performance in the world model matches task performance in the real world.

\subsubsection{Repeated Positioning Accuracy}
\label{subsubsec:repeated acc}
We first test repeated target realization, namely whether a target specified in the world model can be reproduced by the physical robot with high precision. In practice, positioning errors are more likely to occur near the boundary of the workspace and close to the robot's motion limits, so we introduce edge-case data to improve coverage of these rare regions.

As shown in Figure~\ref{fig:relocate}, without edge-case data, the model exhibits clear positioning errors in several boundary regions, and the realized end-effector position deviates from the target specified in the world model. After augmentation, these errors are substantially reduced, and physical execution becomes much better aligned with the target. Table~\ref{tbl:edge_case_ablation} further shows that alignment improves as spatial coverage increases. Notably, adding edge-case data at only about 30\% of the regular-scene data volume already yields clear gains. This result highlights the importance of boundary coverage for accurate world-to-real alignment.

\subsubsection{Interpolation within the Covered Workspace}

Accurate reproduction of a small set of observed points is not enough. A useful interactive world model should also support stable interpolation within the covered workspace. In other words, once the training data covers the operation region at a sufficient level, the model should not only memorize a sparse set of sampled points. It should also produce reasonable and stable interactions at interior points and states that were not explicitly collected.

We evaluate this property with a continuous ten-minute interaction sequence on Push-T, which is showcased in our demonstration. Across the sequence, the rollout traverses many intermediate states not explicitly collected in training, yet the generated interaction remains smooth, with consistent robot and object motion throughout. This provides qualitative evidence that the model supports interpolation within the covered workspace rather than merely memorizing a sparse set of anchor states, while also remaining stable over long interaction horizons.

\subsubsection{Task Level Evaluation Consistency}

\begin{figure}
    \centering
    \includegraphics[width=1\linewidth]{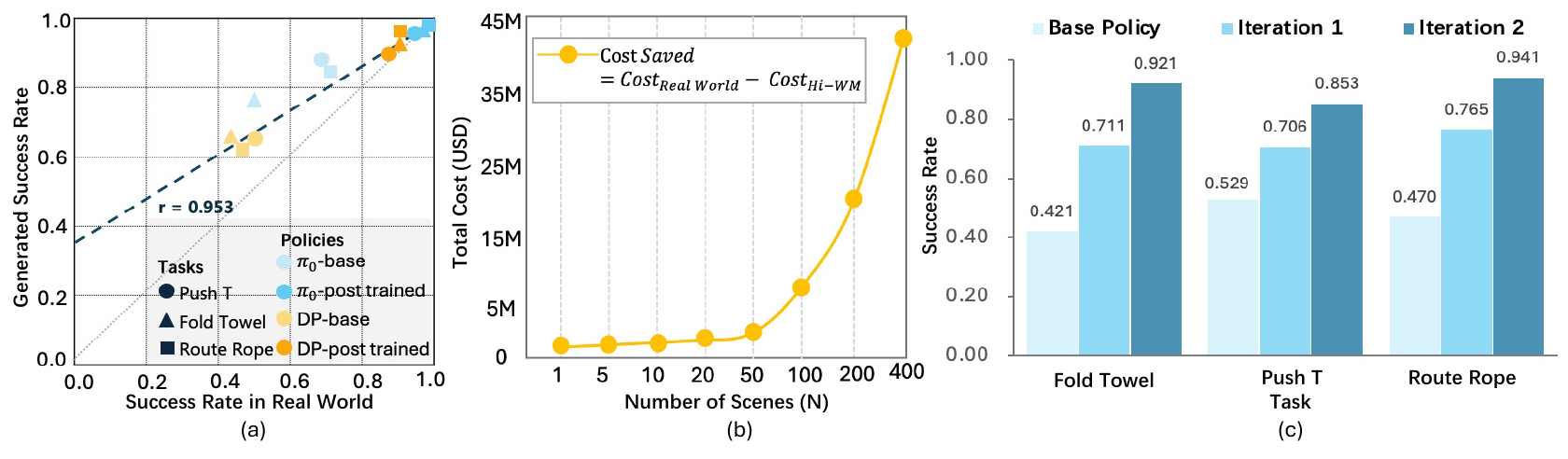}
    \vspace{-1em}
    \caption{\textbf{Correlation, cost savings, and scaling with intervention data.} \textbf{(a):} correlation between success rates measured in the world model and in the real world, showing that world-model evaluation is highly consistent with real execution. \textbf{(b):} cumulative cost saved by conducting human interventions in the world model rather than the real world, demonstrating a substantial and expanding economic advantage as scene coverage increases. \textbf{(c):} scaling of real-world success rate as more intervention data are collected, showing consistent performance improvement across tasks with additional corrective data.}
    \label{fig:corr_cost_scaling}
\end{figure}

Finally, we test whether task performance in the world model correlates with task performance in the real world. As shown in Figure~\ref{fig:corr_cost_scaling}(a), the two success rates are strongly correlated, with a Pearson correlation coefficient of $r = 0.953$.

This result shows that the world model is not only aligned with real world at the execution level, but also predictive at the task level. It can therefore serve as a reliable proxy for policy evaluation. This gives human operators a useful signal for deciding when and where intervention is needed, and it supports the generation of high-quality intervention data for post-training.

\subsection{Policy Performance via Hi-WM}

\begin{table*}[t]
    \centering
    \begin{minipage}[t]{0.46\textwidth}
        \centering
        \centering
\small 
\setlength{\tabcolsep}{4pt}
\renewcommand{\arraystretch}{1.1} 

\caption{
\textbf{Impact of edge-case data scale on visual fidelity.} We measure image prediction quality with PSNR ($\uparrow$), SSIM ($\uparrow$), and LPIPS ($\downarrow$). Percentages indicate the fraction of edge-case trajectories added for augmentation relative to the complete edge-case dataset. Increasing the amount of edge-case data consistently yields higher PSNR and SSIM and lower LPIPS. 
The full augmentation setting achieves the strongest overall fidelity.
}
\label{tbl:edge_case_ablation}

\begin{tabular}{l ccc}
\toprule
\textbf{Ratio (\%)} & \textbf{PSNR} $\uparrow$ & \textbf{SSIM} $\uparrow$ & \textbf{LPIPS} $\downarrow$ \\
\midrule
0 (Base)  & 18.50 & 0.815 & 0.152 \\
20        & 19.82 & 0.853 & 0.124 \\
50        & 21.15 & 0.901 & 0.098 \\
100 (Full)& \textbf{22.53} & \textbf{0.942} & \textbf{0.055} \\
\bottomrule
\end{tabular}
    \end{minipage}
    \hfill
    \begin{minipage}[t]{0.52\textwidth}
        \centering
        \centering
\small 
\setlength{\tabcolsep}{2.5pt}
\renewcommand{\arraystretch}{0.92} 

\caption{
\textbf{Real-world success rates (\%) under the standard task setting.}
Across two policies (DP and $\pi_0$) and three tasks, our Human-in-the-World-Model (Hi-WM) achieves the best result in all six policy--task pairs under matched post-training budgets, outperforming both the base policy and the WM-Closed-Loop (WM-CL) baseline. Best results are in bold.
}

\label{tbl:realworld_wm_success_rate}
\begin{tabular}{ll ccc}
\toprule
\multirow{2}{*}{\textbf{Method}} & \multirow{2}{*}{\textbf{Policy}} & \multicolumn{3}{c}{\textbf{Success Rate (\%)}} \\
\cmidrule(lr){3-5}
& & Fold Towel & Push T & Route Rope \\
\midrule
\multirow{2}{*}{Base} 
& DP  & 42.1 & 52.9 & 47.0 \\
& Pi0 & 55.3 & 76.5 & 64.7 \\
\midrule
\multirow{2}{*}{WM-CL} 
& DP  & 76.3 & 64.7 & 70.6 \\
& Pi0 & 78.9 & 79.4 & 82.4 \\
\midrule
\multirow{2}{*}{\textbf{Ours}} 
& DP  & \textbf{92.1} & \textbf{85.3} & \textbf{94.1} \\
& Pi0 & \textbf{97.4} & \textbf{97.1} & \textbf{100.0}\\
\bottomrule
\end{tabular}
    \end{minipage}
\end{table*}

To answer \textbf{RQ2}, we evaluate whether the resulting corrective trajectories improve downstream policy performance. Table~2 reports real-world success rates on the three tasks under matched post-training budgets for two base policies, DP\cite{diffusion-policy} and $\pi_0$\cite{pi0}. Hi-WM achieves the best result in all six task--policy settings, showing that its benefit is not tied to a particular policy architecture. Averaged across settings, our method improves success by 37.9 percentage points over the base policy and by 19.0 points over the world-model closed-loop baseline. The closed-loop baseline is informative because it isolates the effect of adding additional successful rollouts generated by the current policy without human correction. Although this baseline improves over the base policy in all six settings, it remains consistently below Hi-WM.

We attribute this gap to the type of supervision collected by the two methods. In the closed-loop baseline, the additional data mainly reinforce behaviors already available to the current policy. In contrast, Hi-WM introduces human correction when the rollout begins to fail, so the collected segments are concentrated around failure-prone states that the base policy does not handle well. Moreover, the interactive world model supports rollback and branching from a pre-failure state, allowing multiple corrective continuations to be collected efficiently from the same rollout. This likely provides richer supervision for recovery behavior, which in turn leads to higher real-world success after post-training.

\subsection{Efficiency and Scaling of Hi-WM}

\textbf{Cost scaling with virtual intervention data.}
We first analyze the economic effect of collecting intervention data in the world model rather than in the real world. We define the cost saved as $\text{cost}{\text{real world}} - \text{cost}{\text{Hi-WM}}$. Our analysis includes both fixed setup costs and time-dependent operating costs, including equipment deployment, human labor, real-world site overhead, and inference compute. As shown in Figure~\ref{fig:corr_cost_scaling}(b), the absolute cost savings increase steadily with the number of scenes ($N$). The gap is smaller at low scene counts and widens as $N$ grows, indicating that Hi-WM has a more favorable marginal cost under increasing scene diversity. Replacing real-world collection with Hi-WM yields nearly \$45,000,000 USD in savings. This trend is consistent with the fact that Hi-WM reduces repeated deployment, scene resets, and continuous hardware occupancy during data collection.

\textbf{Scaling with virtual intervention data.}
We next study how real-world policy performance changes as more intervention data are collected in the world model. Starting from the same base policy, we add human corrective data collected through Hi-WM in two stages. Specifically, Iteration 1 augments the training set with virtual intervention data amounting to roughly 20\% of the original offline dataset size, and Iteration 2 increases this cumulative budget to approximately 35\%. We fine-tune the policy after each stage and evaluate the resulting policy in the real world. As shown in Figure~\ref{fig:corr_cost_scaling}(c), success rates improve monotonically from the Base Policy to Iteration 1 and again to Iteration 2 across all three tasks. These results indicate that additional corrective data collected in the world model remain beneficial within the tested range and can translate into consistent gains in real-world task success.

\begin{figure}[t!]
    \centering
    \includegraphics[width=\linewidth]{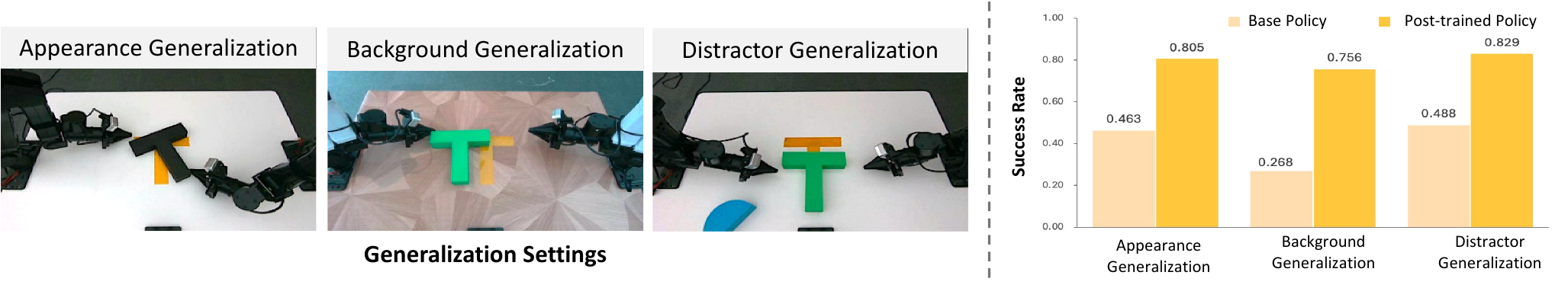}
    \caption{\textbf{Real-world generalization improvements via post-training.} \textbf{Left:} Visualization of diverse real-world generalization settings for Push T, including Appearance Generalization, Background Generalization and Distractor Generalization. \textbf{Right:} Quantitative comparison of success rates. The Post-trained Policy, refined using corrective data collected within the Hi-WM, demonstrates substantial performance improvements over the Base Policy across all evaluated generalization scenarios.}
    \label{fig:general_setting}
\end{figure}

\subsection{Policy Generalization via Hi-WM}
To answer \textbf{RQ4}, we evaluate whether corrective data collected in the world model can improve real-world generalization under held-out scene variations. We focus on Push-T and consider three controlled settings: appearance variation, background variation, and distractor variation. A key advantage of our setting is that the world model can be initialized from varied initial observations under these conditions and rolled out accordingly. This allows human operators to intervene not only on states covered by the original real-world dataset, but also on states induced by these scene variations. We therefore collect corrective data under these varied conditions in the world model, use them for post-training, and evaluate the resulting policy in the real world.

Figure~\ref{fig:general_setting} presents qualitative examples of the three settings and a quantitative comparison between the Base Policy and the post-trained policy. The post-trained policy consistently outperforms the Base Policy across all three evaluated settings, indicating that corrective data collected in the world model transfer to improved robustness under these held-out test conditions. These results suggest that Hi-WM improves policy robustness to held-out appearance, background, and distractor variations in the real world.

\section{Conclusion}
In this paper, we introduced Human-in-the-World-Model (Hi-WM), a novel framework designed to overcome the scalability bottlenecks inherent in the post-training of generalist robot policies. While conventional post-training methods are fundamentally constrained by the high marginal costs of physical deployment, scene resets, and hardware occupancy , Hi-WM shifts the human intervention loop into an interactive world simulator. By enabling mechanisms such as state caching, trajectory branching, and reset-free data collection, our approach transforms corrective data acquisition from a hardware-bound procedure into an amortized interactive process.

Our evaluation across real-world manipulation tasks, encompassing both rigid and deformable objects, demonstrates that interactive targeted correction within the world model effectively improves real-world policy success rates. Ultimately, Hi-WM presents a practical and economically efficient paradigm for scaling post-training. By establishing the world model as the primary workspace for high-frequency evaluation and correction, this framework allows expensive real-world deployments to be reserved for sparse but critical final validation.

\clearpage
\bibliographystyle{plainnat}
\bibliography{main}

\begin{thebibliography}{61}
\providecommand{\natexlab}[1]{#1}
\providecommand{\url}[1]{\texttt{#1}}
\expandafter\ifx\csname urlstyle\endcsname\relax
  \providecommand{\doi}[1]{doi: #1}\else
  \providecommand{\doi}{doi: \begingroup \urlstyle{rm}\Url}\fi

\bibitem[{1X Technologies}(2025)]{1xWorldModel}
{1X Technologies}.
\newblock {1X World Model --- 1x.tech}.
\newblock \url{https://www.1x.tech/discover/1x-world-model}, 2025.
\newblock [Accessed 16-05-2025].

\bibitem[Ali et~al.(2025)Ali, Bai, Bala, Balaji, Blakeman, Cai, Cao, Cao, Cha, Chao, et~al.]{ali2025world}
Arslan Ali, Junjie Bai, Maciej Bala, Yogesh Balaji, Aaron Blakeman, Tiffany Cai, Jiaxin Cao, Tianshi Cao, Elizabeth Cha, Yu-Wei Chao, et~al.
\newblock World simulation with video foundation models for physical ai.
\newblock \emph{arXiv preprint arXiv:2511.00062}, 2025.

\bibitem[Awiszus et~al.(2021)Awiszus, Schubert, and Rosenhahn]{awiszus2021worldgen}
Maren Awiszus, Frederik Schubert, and Bodo Rosenhahn.
\newblock World-gan: a generative model for minecraft worlds.
\newblock In \emph{2021 IEEE Conference on Games (CoG)}, pages 1--8. IEEE, 2021.

\bibitem[Bar et~al.(2024)Bar, Zhou, Tran, Darrell, and LeCun]{bar2024navigation}
Amir Bar, Gaoyue Zhou, Danny Tran, Trevor Darrell, and Yann LeCun.
\newblock Navigation world models.
\newblock \emph{arXiv preprint arXiv:2412.03572}, 2024.

\bibitem[Bharadhwaj et~al.(2024)Bharadhwaj, Dwibedi, Gupta, Tulsiani, Doersch, Xiao, Shah, Xia, Sadigh, and Kirmani]{bharadhwaj2024gen2act}
Homanga Bharadhwaj, Debidatta Dwibedi, Abhinav Gupta, Shubham Tulsiani, Carl Doersch, Ted Xiao, Dhruv Shah, Fei Xia, Dorsa Sadigh, and Sean Kirmani.
\newblock Gen2act: Human video generation in novel scenarios enables generalizable robot manipulation.
\newblock \emph{arXiv preprint arXiv:2409.16283}, 2024.

\bibitem[Black et~al.(2023)Black, Nakamoto, Atreya, Walke, Finn, Kumar, and Levine]{black2023zero}
Kevin Black, Mitsuhiko Nakamoto, Pranav Atreya, Homer Walke, Chelsea Finn, Aviral Kumar, and Sergey Levine.
\newblock Zero-shot robotic manipulation with pretrained image-editing diffusion models.
\newblock \emph{arXiv preprint arXiv:2310.10639}, 2023.

\bibitem[Black et~al.(2024)Black, Brown, Driess, Esmail, Equi, Finn, Fusai, Groom, Hausman, Ichter, Jakubczak, Jones, Ke, Levine, Li-Bell, Mothukuri, Nair, Pertsch, Shi, Tanner, Vuong, Walling, Wang, and Zhilinsky]{pi0}
Kevin Black, Noah Brown, Danny Driess, Adnan Esmail, Michael Equi, Chelsea Finn, Niccolo Fusai, Lachy Groom, Karol Hausman, Brian Ichter, Szymon Jakubczak, Tim Jones, Liyiming Ke, Sergey Levine, Adrian Li-Bell, Mohith Mothukuri, Suraj Nair, Karl Pertsch, Lucy~Xiaoyang Shi, James Tanner, Quan Vuong, Anna Walling, Haohuan Wang, and Ury Zhilinsky.
\newblock $\pi_0$: A vision-language-action flow model for general robot control, 2024.
\newblock URL \url{https://arxiv.org/abs/2410.24164}.

\bibitem[Bruce et~al.(2024)Bruce, Dennis, Edwards, Parker-Holder, Shi, Hughes, Lai, Mavalankar, Steigerwald, Apps, et~al.]{bruce2024genie}
Jake Bruce, Michael~D Dennis, Ashley Edwards, Jack Parker-Holder, Yuge Shi, Edward Hughes, Matthew Lai, Aditi Mavalankar, Richie Steigerwald, Chris Apps, et~al.
\newblock Genie: Generative interactive environments.
\newblock In \emph{Forty-first International Conference on Machine Learning}, 2024.

\bibitem[Chen et~al.(2024)Chen, Guo, He, Zhang, Zhang, Yang, Zhao, and Bian]{chen2024igor}
Xiaoyu Chen, Junliang Guo, Tianyu He, Chuheng Zhang, Pushi Zhang, Derek~Cathera Yang, Li~Zhao, and Jiang Bian.
\newblock Igor: Image-goal representations are the atomic control units for foundation models in embodied ai.
\newblock \emph{arXiv preprint arXiv:2411.00785}, 2024.

\bibitem[Chi et~al.(2023)Chi, Feng, Du, Xu, Cousineau, Burchfiel, and Song]{diffusion-policy}
Cheng Chi, Siyuan Feng, Yilun Du, Zhenjia Xu, Eric Cousineau, Benjamin Burchfiel, and Shuran Song.
\newblock Diffusion policy: Visuomotor policy learning via action diffusion.
\newblock \emph{arXiv preprint arXiv:2303.04137}, 2023.

\bibitem[Chi et~al.(2025)Chi, Jia, Fan, Ju, Mi, Zhang, Qin, Tian, Ge, Li, et~al.]{chi2025wow}
Xiaowei Chi, Peidong Jia, Chun-Kai Fan, Xiaozhu Ju, Weishi Mi, Kevin Zhang, Zhiyuan Qin, Wanxin Tian, Kuangzhi Ge, Hao Li, et~al.
\newblock Wow: Towards a world omniscient world model through embodied interaction.
\newblock \emph{arXiv preprint arXiv:2509.22642}, 2025.

\bibitem[Du et~al.(2023)Du, Yang, Dai, Dai, Nachum, Tenenbaum, Schuurmans, and Abbeel]{du2023learning}
Yilun Du, Sherry Yang, Bo~Dai, Hanjun Dai, Ofir Nachum, Josh Tenenbaum, Dale Schuurmans, and Pieter Abbeel.
\newblock Learning universal policies via text-guided video generation.
\newblock \emph{Advances in neural information processing systems}, 36:\penalty0 9156--9172, 2023.

\bibitem[Gao et~al.(2025)Gao, Belkhale, Dasari, Balakrishna, Shah, and Sadigh]{gao2025stargen}
Jensen Gao, Suneel Belkhale, Sudeep Dasari, Ashwin Balakrishna, Dhruv Shah, and Dorsa Sadigh.
\newblock A taxonomy for evaluating generalist robot manipulation policies.
\newblock \emph{arXiv preprint arXiv:2503.01238}, 2025.

\bibitem[Guo et~al.(2024)Guo, Hu, Zhang, Wang, Chen, Lu, and Chen]{guo2024prediction}
Yanjiang Guo, Yucheng Hu, Jianke Zhang, Yen-Jen Wang, Xiaoyu Chen, Chaochao Lu, and Jianyu Chen.
\newblock Prediction with action: Visual policy learning via joint denoising process.
\newblock In \emph{The Thirty-eighth Annual Conference on Neural Information Processing Systems}, 2024.

\bibitem[Guo et~al.(2025)Guo, Shi, Chen, and Finn]{guo2025ctrl}
Yanjiang Guo, Lucy~Xiaoyang Shi, Jianyu Chen, and Chelsea Finn.
\newblock Ctrl-world: A controllable generative world model for robot manipulation.
\newblock \emph{arXiv preprint arXiv:2510.10125}, 2025.

\bibitem[Guo et~al.(2026)Guo, Lee, Shi, Chen, Liang, and Finn]{guo2026vlaw}
Yanjiang Guo, Tony Lee, Lucy~Xiaoyang Shi, Jianyu Chen, Percy Liang, and Chelsea Finn.
\newblock Vlaw: Iterative co-improvement of vision-language-action policy and world model.
\newblock \emph{arXiv preprint arXiv:2602.12063}, 2026.

\bibitem[HO et~al.(2025)HO, MONAS, REN, and YU]{ho20251x}
D~HO, J~MONAS, JT~REN, and C~YU.
\newblock 1x world model: evaluating bits, not atoms, 2025.

\bibitem[Hoque et~al.(2022)Hoque, Balakrishna, Novoseller, Wilcox, Brown, and Goldberg]{hoque2022thriftydagger}
Ryan Hoque, Ashwin Balakrishna, Ellen Novoseller, Albert Wilcox, Daniel~S. Brown, and Ken Goldberg.
\newblock {ThriftyDAgger}: Budget-aware novelty and risk gating for interactive imitation learning.
\newblock In \emph{Proceedings of the 5th Conference on Robot Learning}, pages 598--608, 2022.

\bibitem[Hoque et~al.(2023)Hoque, Chen, Sharma, Dharmarajan, Thananjeyan, Abbeel, and Goldberg]{hoque2023fleetdagger}
Ryan Hoque, Lawrence~Yunliang Chen, Satvik Sharma, Karthik Dharmarajan, Brijen Thananjeyan, Pieter Abbeel, and Ken Goldberg.
\newblock {Fleet-DAgger}: Interactive robot fleet learning with scalable human supervision.
\newblock In \emph{Proceedings of The 6th Conference on Robot Learning}, pages 368--380, 2023.

\bibitem[Hu et~al.(2024)Hu, Guo, Wang, Chen, Wang, Zhang, Sreenath, Lu, and Chen]{hu2024video}
Yucheng Hu, Yanjiang Guo, Pengchao Wang, Xiaoyu Chen, Yen-Jen Wang, Jianke Zhang, Koushil Sreenath, Chaochao Lu, and Jianyu Chen.
\newblock Video prediction policy: A generalist robot policy with predictive visual representations.
\newblock \emph{arXiv preprint arXiv:2412.14803}, 2024.

\bibitem[Huang et~al.(2025)Huang, Chen, Zhou, Chen, Jiang, Hu, Liao, Gao, Li, Yao, et~al.]{huang2025enerverse}
Siyuan Huang, Liliang Chen, Pengfei Zhou, Shengcong Chen, Zhengkai Jiang, Yue Hu, Yue Liao, Peng Gao, Hongsheng Li, Maoqing Yao, et~al.
\newblock Enerverse: Envisioning embodied future space for robotics manipulation.
\newblock \emph{arXiv preprint arXiv:2501.01895}, 2025.

\bibitem[Jang et~al.(2025)Jang, Ye, Lin, Xiang, Bjorck, Fang, Hu, Huang, Kundalia, Lin, Magne, Mandlekar, Narayan, Tan, Wang, Wang, Wang, Xu, Zeng, Zheng, Zheng, Liu, Zettlemoyer, Fox, Kautz, Reed, Zhu, and Fan]{jang2025dreamgen}
Joel Jang, Seonghyeon Ye, Zongyu Lin, Jiannan Xiang, Johan Bjorck, Yu~Fang, Fengyuan Hu, Spencer Huang, Kaushil Kundalia, Yen-Chen Lin, Lo{\"i}c Magne, Ajay Mandlekar, Avnish Narayan, You~Liang Tan, Guanzhi Wang, Jing Wang, Qi~Wang, Yinzhen Xu, Xiaohui Zeng, Kaiyuan Zheng, Ruijie Zheng, Ming-Yu Liu, Luke Zettlemoyer, Dieter Fox, Jan Kautz, Scott Reed, Yuke Zhu, and Linxi Fan.
\newblock {DreamGen}: Unlocking generalization in robot learning through video world models.
\newblock In \emph{Proceedings of The 9th Conference on Robot Learning}, pages 5170--5194, 2025.

\bibitem[Jauhri et~al.(2021)Jauhri, Celemin, and Kober]{jauhri2021tips}
Snehal Jauhri, Carlos Celemin, and Jens Kober.
\newblock Interactive imitation learning in state-space.
\newblock In \emph{Proceedings of the 2020 Conference on Robot Learning}, pages 682--692, 2021.

\bibitem[Jia et~al.(2025)Jia, Liu, Liu, Zhou, Yu, Yan, Chi, Guo, Shi, and Zhang]{jia2025video2act}
Yueru Jia, Jiaming Liu, Shengbang Liu, Rui Zhou, Wanhe Yu, Yuyang Yan, Xiaowei Chi, Yandong Guo, Boxin Shi, and Shanghang Zhang.
\newblock Video2act: A dual-system video diffusion policy with robotic spatio-motional modeling.
\newblock \emph{arXiv preprint arXiv:2512.03044}, 2025.

\bibitem[Jiang et~al.(2025{\natexlab{a}})Jiang, Wang, Zhang, Wu, and Li]{jiang2024transic}
Yunfan Jiang, Chen Wang, Ruohan Zhang, Jiajun Wu, and Fei-Fei Li.
\newblock {TRANSIC}: Sim-to-real policy transfer by learning from online correction.
\newblock In \emph{Proceedings of The 8th Conference on Robot Learning}, pages 1691--1729, 2025{\natexlab{a}}.

\bibitem[Jiang et~al.(2025{\natexlab{b}})Jiang, Chen, Huang, Chen, Zhou, Liao, He, Liu, Li, Yao, et~al.]{jiang2025enerverse}
Yuxin Jiang, Shengcong Chen, Siyuan Huang, Liliang Chen, Pengfei Zhou, Yue Liao, Xindong He, Chiming Liu, Hongsheng Li, Maoqing Yao, et~al.
\newblock Enerverse-ac: Envisioning embodied environments with action condition.
\newblock \emph{arXiv preprint arXiv:2505.09723}, 2025{\natexlab{b}}.

\bibitem[Jiang et~al.(2025{\natexlab{c}})Jiang, Liu, Qin, Tian, Zheng, Zhou, Yu, Li, and Zhao]{jiang2025world4rl}
Zhennan Jiang, Kai Liu, Yuxin Qin, Shuai Tian, Yupeng Zheng, Mingcai Zhou, Chao Yu, Haoran Li, and Dongbin Zhao.
\newblock World4rl: Diffusion world models for policy refinement with reinforcement learning for robotic manipulation.
\newblock \emph{arXiv preprint arXiv:2509.19080}, 2025{\natexlab{c}}.

\bibitem[Jiang et~al.(2026)Jiang, Zhou, Jiang, Huang, Wei, Chen, Zhou, Guo, Lin, Zhang, et~al.]{jiang2026wovr}
Zhennan Jiang, Shangqing Zhou, Yutong Jiang, Zefang Huang, Mingjie Wei, Yuhui Chen, Tianxing Zhou, Zhen Guo, Hao Lin, Quanlu Zhang, et~al.
\newblock Wovr: World models as reliable simulators for post-training vla policies with rl.
\newblock \emph{arXiv preprint arXiv:2602.13977}, 2026.

\bibitem[Kelly et~al.(2019)Kelly, Sidrane, Driggs-Campbell, and Kochenderfer]{kelly2019hgdagger}
Michael Kelly, Chelsea Sidrane, Katherine Driggs-Campbell, and Mykel~J. Kochenderfer.
\newblock {HG-DAgger}: Interactive imitation learning with human experts.
\newblock In \emph{2019 International Conference on Robotics and Automation (ICRA)}, pages 8077--8083, 2019.

\bibitem[Koh et~al.(2021)Koh, Lee, Yang, Baldridge, and Anderson]{koh2021pathdreamer}
Jing~Yu Koh, Honglak Lee, Yinfei Yang, Jason Baldridge, and Peter Anderson.
\newblock Pathdreamer: A world model for indoor navigation.
\newblock In \emph{Proceedings of the IEEE/CVF International Conference on Computer Vision}, pages 14738--14748, 2021.

\bibitem[Laskey et~al.(2017)Laskey, Lee, Fox, Dragan, and Goldberg]{laskey2017dart}
Michael Laskey, Jonathan Lee, Roy Fox, Anca Dragan, and Ken Goldberg.
\newblock {DART}: Noise injection for robust imitation learning.
\newblock In \emph{Proceedings of the 1st Annual Conference on Robot Learning}, pages 143--156, 2017.

\bibitem[Li et~al.(2025{\natexlab{a}})Li, Krause, and Hutter]{li2025robotic}
Chenhao Li, Andreas Krause, and Marco Hutter.
\newblock Robotic world model: A neural network simulator for robust policy optimization in robotics.
\newblock \emph{arXiv preprint arXiv:2501.10100}, 2025{\natexlab{a}}.

\bibitem[Li et~al.(2025{\natexlab{b}})Li, Hsu, Gu, Mees, Pertsch, Walke, Fu, Lunawat, Sieh, Kirmani, Levine, Wu, Finn, Su, Vuong, and Xiao]{li2025simpler}
Xuanlin Li, Kyle Hsu, Jiayuan Gu, Oier Mees, Karl Pertsch, Homer~Rich Walke, Chuyuan Fu, Ishikaa Lunawat, Isabel Sieh, Sean Kirmani, Sergey Levine, Jiajun Wu, Chelsea Finn, Hao Su, Quan Vuong, and Ted Xiao.
\newblock Evaluating real-world robot manipulation policies in simulation.
\newblock In \emph{Proceedings of The 8th Conference on Robot Learning}, pages 3705--3728, 2025{\natexlab{b}}.

\bibitem[Li et~al.(2025{\natexlab{c}})Li, Zhu, Wen, Shen, and Xu]{li2025worldeval}
Yaxuan Li, Yichen Zhu, Junjie Wen, Chaomin Shen, and Yi~Xu.
\newblock Worldeval: World model as real-world robot policies evaluator.
\newblock \emph{arXiv preprint arXiv:2505.19017}, 2025{\natexlab{c}}.

\bibitem[Liao et~al.(2025)Liao, Zhou, Huang, Yang, Chen, Jiang, Hu, Cai, Liu, Luo, et~al.]{liao2025genie}
Yue Liao, Pengfei Zhou, Siyuan Huang, Donglin Yang, Shengcong Chen, Yuxin Jiang, Yue Hu, Jingbin Cai, Si~Liu, Jianlan Luo, et~al.
\newblock Genie envisioner: A unified world foundation platform for robotic manipulation.
\newblock \emph{arXiv preprint arXiv:2508.05635}, 2025.

\bibitem[Liu et~al.(2023)Liu, Nasiriany, Zhang, Bao, and Zhu]{liu2023sirius}
Huihan Liu, Soroush Nasiriany, Lance Zhang, Zhiyao Bao, and Yuke Zhu.
\newblock Robot learning on the job: Human-in-the-loop autonomy and learning during deployment.
\newblock In \emph{Robotics: Science and Systems (RSS)}, 2023.

\bibitem[Liu et~al.(2026)Liu, Bai, Ci, Ma, and Shou]{liu2026world}
Xiaokang Liu, Zechen Bai, Hai Ci, Kevin~Yuchen Ma, and Mike~Zheng Shou.
\newblock World-vla-loop: Closed-loop learning of video world model and vla policy.
\newblock \emph{arXiv preprint arXiv:2602.06508}, 2026.

\bibitem[Luo et~al.(2024{\natexlab{a}})Luo, Hu, Xu, Tan, Berg, Sharma, Schaal, Finn, Gupta, and Levine]{luo2024serl}
Jianlan Luo, Zheyuan Hu, Charles Xu, You~Liang Tan, Jacob Berg, Archit Sharma, Stefan Schaal, Chelsea Finn, Abhishek Gupta, and Sergey Levine.
\newblock Serl: A software suite for sample-efficient robotic reinforcement learning, 2024{\natexlab{a}}.

\bibitem[Luo et~al.(2024{\natexlab{b}})Luo, Xu, Wu, and Levine]{luo2024hilserl}
Jianlan Luo, Charles Xu, Jeffrey Wu, and Sergey Levine.
\newblock Precise and dexterous robotic manipulation via human-in-the-loop reinforcement learning, 2024{\natexlab{b}}.

\bibitem[Mandlekar et~al.(2023)Mandlekar, Garrett, Xu, and Fox]{Mandlekar2023}
Ajay Mandlekar, Caelan Garrett, Danfei Xu, and Dieter Fox.
\newblock Human-in-the-loop task and motion planning for imitation learning.
\newblock In \emph{Conference on Robot Learning (CoRL)}, volume 229 of \emph{PMLR}, 2023.

\bibitem[Mendonca et~al.(2023)Mendonca, Bahl, and Pathak]{mendonca2023structuredwm}
Russell Mendonca, Shikhar Bahl, and Deepak Pathak.
\newblock Structured world models from human videos.
\newblock In \emph{Robotics: Science and Systems (RSS)}, 2023.

\bibitem[Qiu et~al.(2025)Qiu, Wang, Cai, Chen, Lin, Wang, and Gan]{qiu2025lucibot}
Xiaowen Qiu, Yian Wang, Jiting Cai, Zhehuan Chen, Chunru Lin, Tsun-Hsuan Wang, and Chuang Gan.
\newblock {LuciBot}: Automated robot policy learning from generated videos.
\newblock \emph{arXiv preprint arXiv:2503.09871}, 2025.

\bibitem[Quevedo et~al.(2025)Quevedo, Sharma, Sun, Suryavanshi, Liang, and Yang]{quevedo2025worldgym}
Julian Quevedo, Ansh~Kumar Sharma, Yixiang Sun, Varad Suryavanshi, Percy Liang, and Sherry Yang.
\newblock {WorldGym}: World model as an environment for policy evaluation.
\newblock \emph{arXiv preprint arXiv:2506.00613}, 2025.

\bibitem[Ren et~al.(2023)Ren, Zhang, Zheng, Li, Wen, Zhu, Ma, and Liang]{ren2023surfer}
Pengzhen Ren, Kaidong Zhang, Hetao Zheng, Zixuan Li, Yuhang Wen, Fengda Zhu, Mas Ma, and Xiaodan Liang.
\newblock Surfer: Progressive reasoning with world models for robotic manipulation.
\newblock \emph{arXiv preprint arXiv:2306.11335}, 2023.

\bibitem[Ross et~al.(2011)Ross, Gordon, and Bagnell]{ross2011dagger}
St{\'e}phane Ross, Geoffrey~J. Gordon, and J.~Andrew Bagnell.
\newblock A reduction of imitation learning and structured prediction to no-regret online learning.
\newblock In \emph{Proceedings of the Fourteenth International Conference on Artificial Intelligence and Statistics}, pages 627--635, 2011.

\bibitem[Seo et~al.(2023{\natexlab{a}})Seo, Kim, James, Lee, Shin, and Abbeel]{seo2023multi}
Younggyo Seo, Junsu Kim, Stephen James, Kimin Lee, Jinwoo Shin, and Pieter Abbeel.
\newblock Multi-view masked world models for visual robotic manipulation.
\newblock In \emph{International Conference on Machine Learning}, pages 30613--30632. PMLR, 2023{\natexlab{a}}.

\bibitem[Seo et~al.(2023{\natexlab{b}})Seo, Kim, James, Lee, Shin, and Abbeel]{seo2023mwm}
Younggyo Seo, Junsu Kim, Stephen James, Kimin Lee, Jinwoo Shin, and Pieter Abbeel.
\newblock Multi-view masked world models for visual robotic manipulation.
\newblock In \emph{Proceedings of the 40th International Conference on Machine Learning}, pages 30613--30632, 2023{\natexlab{b}}.

\bibitem[Sharma et~al.(2026)Sharma, Sun, Lu, Zhang, Liu, and Yang]{sharma2026world}
Ansh~Kumar Sharma, Yixiang Sun, Ninghao Lu, Yunzhe Zhang, Jiarao Liu, and Sherry Yang.
\newblock World-gymnast: Training robots with reinforcement learning in a world model.
\newblock \emph{arXiv preprint arXiv:2602.02454}, 2026.

\bibitem[Spencer et~al.(2020)Spencer, Choudhury, Barnes, Schmittle, Chiang, Ramadge, and Srinivasa]{spencer2020interventions}
Jonathan Spencer, Sanjiban Choudhury, Matt Barnes, Matthew Schmittle, Mung Chiang, Peter Ramadge, and Siddhartha Srinivasa.
\newblock Learning from interventions: Human-robot interaction as both explicit and implicit feedback.
\newblock In \emph{Proceedings of Robotics: Science and Systems}, 2020.
\newblock \doi{10.15607/RSS.2020.XVI.055}.

\bibitem[Team et~al.(2025)Team, Devin, Du, Dwibedi, Gao, Jindal, Kipf, Kirmani, Liu, Majumdar, et~al.]{team2025evaluating}
Gemini~Robotics Team, Coline Devin, Yilun Du, Debidatta Dwibedi, Ruiqi Gao, Abhishek Jindal, Thomas Kipf, Sean Kirmani, Fangchen Liu, Anirudha Majumdar, et~al.
\newblock Evaluating gemini robotics policies in a veo world simulator.
\newblock \emph{arXiv preprint arXiv:2512.10675}, 2025.

\bibitem[Wang et~al.(2026)]{wang2026interactive}
Yixuan Wang et~al.
\newblock Interactive world simulator for robot policy training and evaluation.
\newblock \emph{arXiv preprint arXiv:2603.08546}, 2026.

\bibitem[Welte et~al.(2025)]{Welte2025}
Elise Welte et~al.
\newblock Interactive imitation learning for dexterous robotic manipulation: Challenges and perspectives.
\newblock \emph{Frontiers in Robotics and AI}, 12, 2025.

\bibitem[Wu et~al.(2023)Wu, Escontrela, Hafner, Abbeel, and Goldberg]{wu2023daydreamer}
Philipp Wu, Alejandro Escontrela, Danijar Hafner, Pieter Abbeel, and Ken Goldberg.
\newblock Daydreamer: World models for physical robot learning.
\newblock In \emph{Conference on robot learning}, pages 2226--2240. PMLR, 2023.

\bibitem[Wu et~al.(2025)Wu, Shentu, Liao, Jin, Guo, Sreenath, Lin, and Abbeel]{wu2025robocopilot}
Philipp Wu, Yide Shentu, Qiayuan Liao, Ding Jin, Menglong Guo, Koushil Sreenath, Xingyu Lin, and Pieter Abbeel.
\newblock {RoboCopilot}: Human-in-the-loop interactive imitation learning for robot manipulation.
\newblock \emph{arXiv preprint arXiv:2503.07771}, 2025.

\bibitem[Xiao et~al.(2025)Xiao, Yang, Chang, Chen, Xiong, Xu, Zheng, and Zhang]{xiao2025world}
Junjin Xiao, Yandan Yang, Xinyuan Chang, Ronghan Chen, Feng Xiong, Mu~Xu, Wei-Shi Zheng, and Qing Zhang.
\newblock World-env: Leveraging world model as a virtual environment for vla post-training.
\newblock \emph{arXiv preprint arXiv:2509.24948}, 2025.

\bibitem[Ye et~al.(2024)Ye, Jang, Jeon, Joo, Yang, Peng, Mandlekar, Tan, Chao, Lin, et~al.]{ye2024latent}
Seonghyeon Ye, Joel Jang, Byeongguk Jeon, Sejune Joo, Jianwei Yang, Baolin Peng, Ajay Mandlekar, Reuben Tan, Yu-Wei Chao, Bill~Yuchen Lin, et~al.
\newblock Latent action pretraining from videos.
\newblock \emph{arXiv preprint arXiv:2410.11758}, 2024.

\bibitem[Zhang et~al.(2024)Zhang, Ren, Lin, Lin, Ma, Xu, and Liang]{zhang2024pivot}
Kaidong Zhang, Pengzhen Ren, Bingqian Lin, Junfan Lin, Shikui Ma, Hang Xu, and Xiaodan Liang.
\newblock Pivot-r: Primitive-driven waypoint-aware world model for robotic manipulation.
\newblock \emph{arXiv preprint arXiv:2410.10394}, 2024.

\bibitem[Zhang et~al.(2026)Zhang, Yang, Lu, Guo, Zhang, Hu, and Chen]{zhang2026veo}
Zhongru Zhang, Chenghan Yang, Qingzhou Lu, Yanjiang Guo, Jianke Zhang, Yucheng Hu, and Jianyu Chen.
\newblock Veo-act: How far can frontier video models advance generalizable robot manipulation?
\newblock \emph{arXiv preprint arXiv:2604.04502}, 2026.

\bibitem[Zhou et~al.(2025{\natexlab{a}})Zhou, Chen, Chen, Chen, Zhao, Jin, Ren, and Luo]{zhou2025act2goal}
Pengfei Zhou, Liliang Chen, Shengcong Chen, Di~Chen, Wenzhi Zhao, Rongjun Jin, Guanghui Ren, and Jianlan Luo.
\newblock Act2goal: From world model to general goal-conditioned policy.
\newblock \emph{arXiv preprint arXiv:2512.23541}, 2025{\natexlab{a}}.

\bibitem[Zhou et~al.(2025{\natexlab{b}})Zhou, Atreya, Tan, Pertsch, and Levine]{zhou2025autoeval}
Zhiyuan Zhou, Pranav Atreya, You~Liang Tan, Karl Pertsch, and Sergey Levine.
\newblock Autoeval: Autonomous evaluation of generalist robot manipulation policies in the real world.
\newblock \emph{arXiv preprint arXiv:2503.24278}, 2025{\natexlab{b}}.

\bibitem[Zhu et~al.(2024)Zhu, Wu, Guo, Liu, Cheang, and Kong]{zhu2024maniwm}
Fangqi Zhu, Hongtao Wu, Song Guo, Yuxiao Liu, Chilam Cheang, and Tao Kong.
\newblock Mani-{WM}: An interactive world model for real-robot manipulation, 2024.
\newblock URL \url{https://openreview.net/forum?id=aVyJwS1fqQ}.

\end{thebibliography}

\clearpage


\end{document}